\documentclass[a4paper]{llncs}
\usepackage{float}         
\usepackage{amssymb}
\usepackage{graphicx}
\usepackage{comment}
\usepackage{soul,color}
\usepackage[caption=false]{subfig}
\usepackage{algorithm}
\usepackage{algorithmic}
\usepackage{tabularx}

\graphicspath{{figures/}}
\usepackage{url}
\urldef{\mailsa}\path|{alfred.hofmann, ursula.barth, ingrid.haas, frank.holzwarth,|
\urldef{\mailsb}\path|anna.kramer, leonie.kunz, christine.reiss, nicole.sator,|
\urldef{\mailsc}\path|erika.siebert-cole, peter.strasser, lncs}@springer.com|    
\newcommand{\keywords}[1]{\par\addvspace\baselineskip
\noindent\keywordname\enspace\ignorespaces#1}

\makeatletter
\renewcommand\section{\@startsection{section}{1}{\z@}%
	{-4\p@ \@plus -4\p@ \@minus -4\p@}%
	{6\p@ \@plus 4\p@ \@minus 4\p@}%
	{\normalfont\large\bfseries\boldmath
		\rightskip=\z@ \@plus 8em\pretolerance=10000 }}
\makeatother
\begin{document}

\mainmatter  

\title{Kid-Net: Convolution Networks for Kidney Vessels Segmentation from CT-Volumes}


%
%
\author{Ahmed Taha \inst{1}
\and Pechin Lo \inst{2}
\and Junning Li \inst{2}
\and Tao Zhao  \inst{2}}
%

\institute{University of Maryland, College Park , ahmdtaha@cs.umd.edu\and
Intuitive Surgical, Inc, \{firstname.lastname\}@intusurg.com} 

%

%
%
\maketitle

\begin{abstract}
Semantic image segmentation plays an important role in modeling patient-specific anatomy. We propose a convolution neural network, called Kid-Net, along with a training schema to segment kidney vessels: artery, vein and collecting system. Such segmentation is vital during the surgical planning phase in which medical decisions are made before surgical incision. Our main contribution is developing a training schema that handles unbalanced data, reduces false positives and enables high-resolution segmentation with a limited memory budget. These objectives are attained using dynamic weighting, random sampling and $3D$ patch segmentation.\\
Manual medical image annotation is both time-consuming and expensive. Kid-Net reduces kidney vessels segmentation time from matter of hours to minutes. It is trained end-to-end using $3D$ patches from volumetric CT-images. A complete segmentation for a $512\times512\times512$ CT-volume is obtained within a few minutes (1-2 mins) by stitching the output $3D$ patches together. Feature down-sampling and up-sampling are utilized to achieve higher classification and localization accuracies. Quantitative and qualitative evaluation results on a challenging testing dataset show Kid-Net competence. 
\end{abstract}

\section{Introduction and Related Work}

{After its success in classification~\cite{krizhevsky2012imagenet} and action recognition~\cite{ji20133d}, convolution neural networks (CNN) began achieving promising results in challenging semantic segmentation tasks~\cite{milletari2016v,ronneberger2015u,bertasius2016semantic}. One key pillar is its ability to learn features from raw input data-- without relying on hand-crafted features. A second recent pillar is the ability to precisely localize these features when combining convolution features at different scales. Such localization approach eliminate the need for traditional hand-crafted post processing like Dense-CRF~\cite{chen2016deeplab,krahenbuhl2011efficient}. Thus, end-to-end CNN training for challenging segmentation problems becomes feasible.  This sheds light on semantic segmentation applications in medical field.

Semantic segmentation for human anatomy using volumetric scans like MRI and CT-volumes is an important medical application. It is a fundamental step to perform or plan surgical procedures. Recent work uses automatic segmentation to do  computer assisted diagnosis~\cite{porter2003combining,shin2016deep}, interventions~\cite{zettinig2015multimodal} and segmentation from sparse annotations~\cite{cciccek20163d}. Recently, U-shaped networks~\cite{ronneberger2015u} managed to train fully $2D$ convolution network for semantic segmentation in an end-to-end fashion. These architectures have two contradicting phases that carry out complementary tasks. The down-sampling phase detects features, while the up-sampling phase accurately localizes the detected features. Such combination is proven essential in recent literature~\cite{merkow2016dense,cciccek20163d,milletari2016v} to acquire precise segmentation.

Inspired by U-shaped networks, we propose Kid-Net to segment $3D$ patches from volumetric CT-images. We agree with~\cite{milletari2016v} that $3D$ convolutions are better than slice-wise approaches when processing volumetric scans. We build on~\cite{milletari2016v} architecture by processing volumetric patches to enable high resolution segmentation and thus bypass GPU memory constraints. Despite the promising results, such vanilla model suffers due to unbalanced data. This leads to our main contribution which is balancing both intra-foreground and background-foreground classes within independent patches. This achieves the best results as presented in the experiments section. Accordingly, manual preprocessing like cropping or down-sampling workarounds are no longer needed for high resolution CT-volume segmentation. 


In this work, we aim to segment kidney vessels: artery, vein and collecting system (ureter). This task is challenging for a number of reasons. First, the CT-volume is huge to fit in memory. To avoid processing the whole CT-volume, we process $3D$ patches individually. Second, foreground and background classes are unbalanced and most patches are foreground-free. Another major challenge is obtaining the groundtruth for training. Medical staff annotate our data; their prior knowledge leads to incomplete groundtruth. Vessels far from kidney are considered less relevant and thus ignored. Figure~\ref{fig:ct_image_segmentation} shows a CT-slice with the three foreground classes. It highlights the problem  difficulty  even for a well-informed technician.

\begin{figure}
	\subfloat[CT Slice]{\includegraphics[width=0.22\textwidth,height=0.25\textwidth]{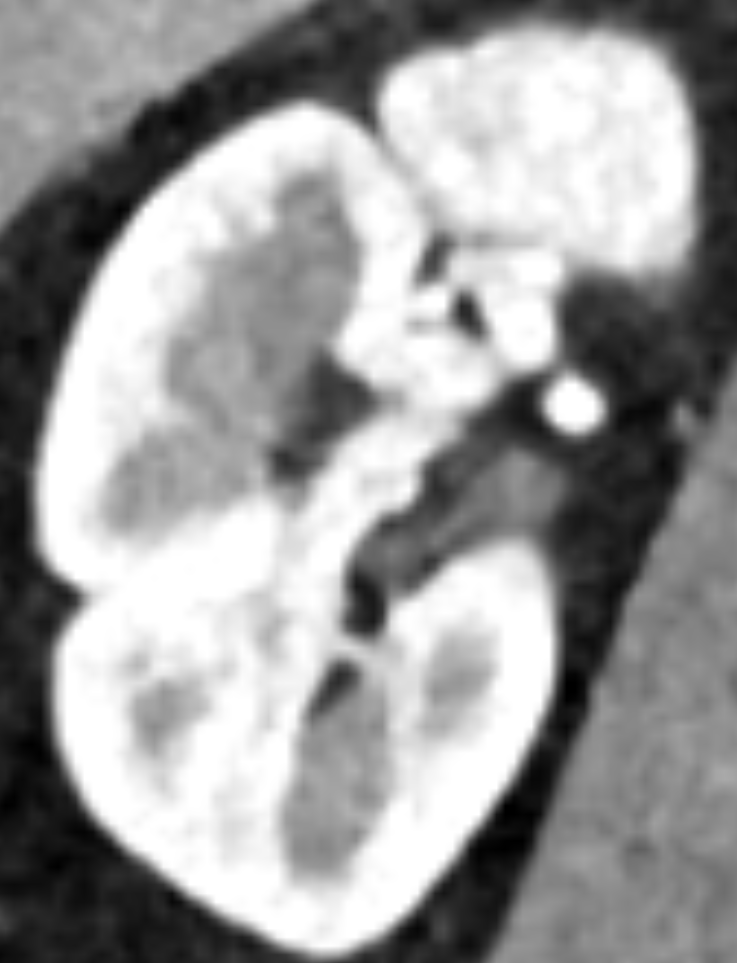}}
	\hfill\subfloat[Artery]{\includegraphics[width=0.22\textwidth,height=0.25\textwidth]{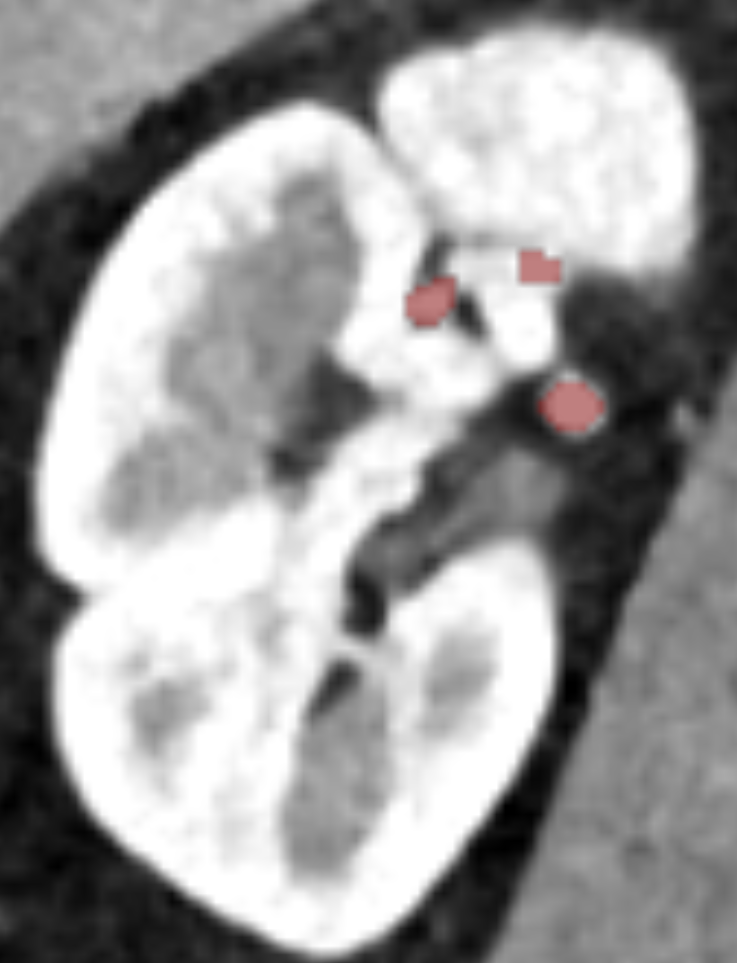}}
	\hfill\subfloat[Vein]{\includegraphics[width=0.22\textwidth,height=0.25\textwidth]{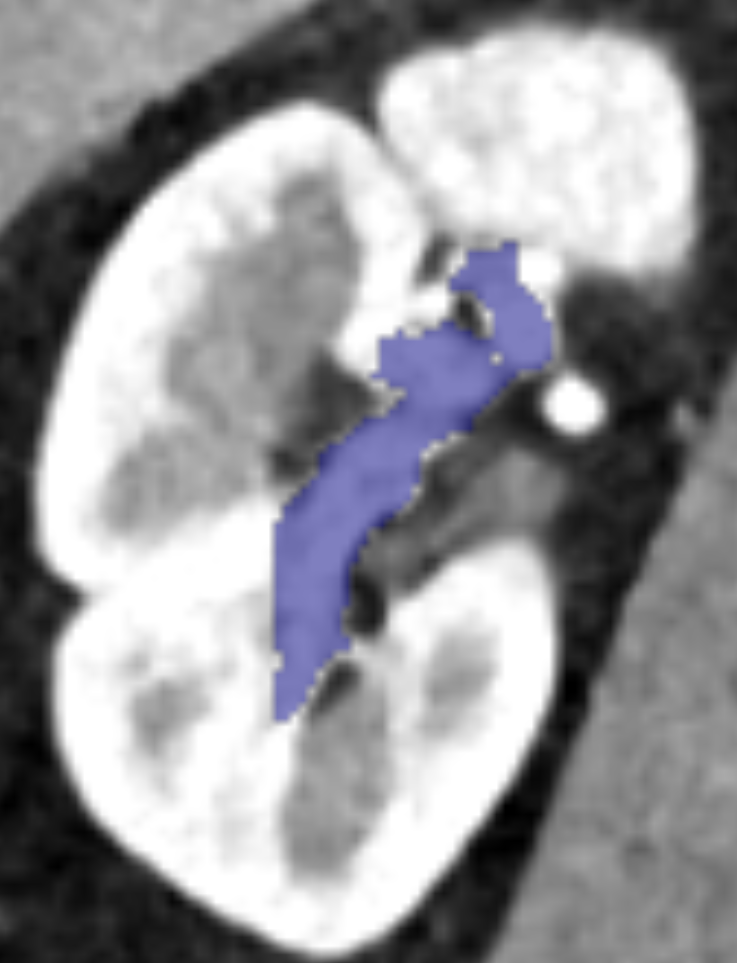}}
	\hfill\subfloat[Ureter]{\includegraphics[width=0.22\textwidth,height=0.25\textwidth]{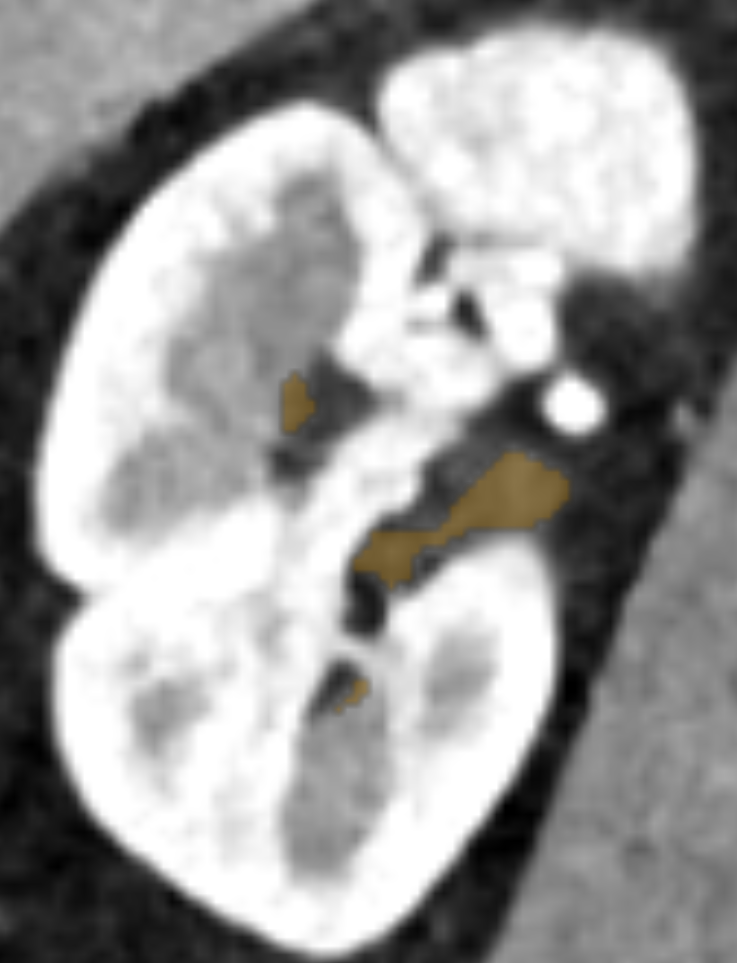}}
	\caption{CT slice contains three foreground classes: artery (red), vein (blue) and collecting system (orange). Best seen in color and zoom}
	\label{fig:ct_image_segmentation}
\end{figure}

\section{Method}

Inspired by U-Net structure~\cite{ronneberger2015u}, Kid-Net is divided into two main phases: down-sampling and up-sampling. Figure~\ref{fig:iris_net} shows these phases and how Kid-Net up-sampling phase is different from U-Net. In U-Net, down-sampled features are repeatedly up-sampled and concatenated with the corresponding feature till a single segmentation result, with the original image resolution, is obtained. Kid-Net is similar to U-Net but adds residual links extension in which each down-sampled feature is independently up-sampled $2^n$ times till the original resolution is restored. Thus unlike U-Net, Kid-Net generates multiple segmentation results that are averaged to obtain a final segmentation result. These residual links follow Junning et. al~\cite{junningresunet2018} Residual-U-Net design. Sequential non-linear functions accumulation improves deep neural networks performance in image recognition. In our paper, residual links are added in up-sampling phase only for simplicity

Kid-Net segments kidney vessels from CT-volumes. The foreground classes are artery, vein and collecting system (ureter) vessels. To avoid the large memory requirement of CT-volumes, Kid-Net is trained using $3D$ CT-volume patches from $R^{96\times96\times96}$. Without any architecture modification, wider context through bigger patches, within GPU memory constraints, are feasible. Kid-Net outputs a soft-max probability maps for artery, vein, collecting system and background. Instead of training for individual foreground classes independently, our network is trained to detect the three foreground classes. Such approach has two advantages; first, a single network decision per voxel fills in for a heuristic-based approach to merge multiple networks decisions. Second, this approach aligns with~\cite{kaiser2017one} recommendation that learning tasks with less data benefit largely from joint training with other tasks.  

\begin{figure}
	\begin{center}
			\includegraphics[width=1.0\textwidth]{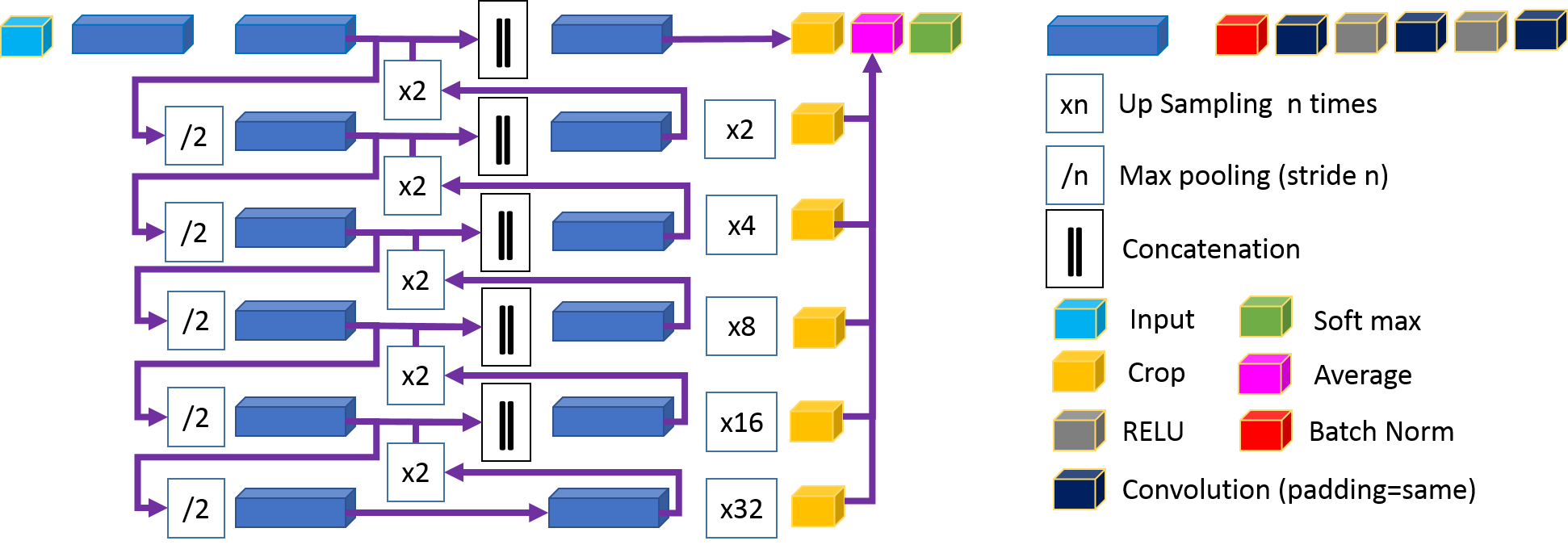}
	\end{center}
	\caption 
	{ \label{fig:iris_net} KID-Net architecture. The two contradicting phases are colored in blue. The down-sampling and up-sampling phases detect and localize features respectively. The segmentation result, at different scale levels, are averaged to compute the final segmentation. Best seen in color and zoom} 
\end{figure}

While training Kid-Net with $3D$ patches bypasses GPU memory constraints, a new challenge surfaces-- unbalanced data.  Patches majority are foreground-free. Even when a foreground class exists, it occupies a small percentage. In this paper, we leverage recent work~\cite{merkow2016dense,cciccek20163d}, that addresses tiny structures precise localization, to tackle unbalanced data. As follows, a two-fold approach is proposed to hinder bias against the tiny structured foreground classes.

The first fold assigns dynamic weights to individual voxels based on their classes and patches significance. A major challenge in medical image processing is detecting small structures. Vessels in direct contact with kidney are tiny and their resolution is limited by acquisition. Tiny vessels are challenging to annotate, more valuable to acquire. Thus, patches containing smaller vessels are more significant, have higher weight. Patch weight is inversely proportional to the vessel volume inside it. Foreground classes are also assigned higher weights to hinder the network background bias. The key idea is to measure the average foreground classes volumes per patch dynamically during training. Then, assign higher weights to classes with smaller average volumes and vice versa.

A policy that a vessel volume $\times$ weight must equal $ 1/n$ is imposed where $n$ is the number of classes including the background class. Thus, all classes contribute equally from a network perspective. To account for data augmentation, vessels volumes are measured during training. Enforcing equal contribution (volume $\times$ weight) from all classes is our objective . To do so, we use the moving average procedure outlined in algorithm~\ref{alg:weighting}.

\begin{algorithm}
		\scriptsize
	\caption{Our proposed moving average procedure assigns voxels dynamic weights ($VW_c$) based on their classes and patch weight. Patch Weight ($PW$) is inversely proportional to its foreground volume. Class weight ($CW$) is inversely proportional to its average volume per patch. Initially $V_c = \frac{1}{n}$ for every class $c$. Our settings $\alpha = 0.001$, $n =4$.}
	\begin{algorithmic}
		\REQUIRE $\alpha:$ Convergence Rate 
		\REQUIRE $P:$ Current $3D$ patch with axis $x,y,z$
		\REQUIRE $n:$ Number of classes (background included) 
	\REQUIRE $V_c:$ Class (c) moving average volume 
		\REQUIRE $PW:$ Current patch weight
		\REQUIRE $CW_c:$ Class (c) weight

		\FORALL{c in classes} 
			\STATE // Measure class (c) volume in patch P
			\STATE $V_c(P) = \left( \sum_{ x}{ \sum_{ y} {\sum_{ z} {P(x,y,z) == c}}} \right) / size(P) $ 
			\STATE // Update class (c) moving average volume
			\STATE $V_c = V_c \times (1-\alpha) +  V_c(P) \times \alpha $ 
			\STATE // Update class weight based on its moving average volume
			\STATE $CW_c = 1 / \left(n \times V_c\right)$
		\ENDFOR
		\STATE // Set patch weight based on foreground volume
		\IF {$P \mbox{ contains background only}$}
			\STATE $PW =1$
		\ELSE
					\STATE // Foreground volume $\sum_{c=1}^{n-1}{V_c(P)} < 1$
			\STATE $PW = 1- log(\sum_{c=1}^{n-1}{V_c(P)}) $ 
		\ENDIF
		\STATE $VW_c = PW * CW_{c} $ (Voxel weight is function of $PW$ and $CW_c$)
	\end{algorithmic}
	\label{alg:weighting}
\end{algorithm}
Due to background relative huge volume, it's assigned tiny class weight. So the network produces a lot of false positives -- it is cheap to mis-classify a background voxel. To tackle this undesired phenomena, we propose our second complementary fold -- Random Sampling. Random Background voxels are sampled and assigned high weights. Such method is most effective in limiting false positives because high loss is incurred for these voxels if mis-classified. Figure~\ref{fig:sampling} shows our sampling schema. Two bands are constructed around kidney vessels using morphological kernel, a binary circular dilation of radii two and four. Misclassifications within the first band ($< 2$ voxels away from the vessel) are  considered marginal errors. In a given patch, the sampled background voxels are equivalent to the foreground vessel volume, where $20\%$ and $80\%$  come from the red band and the volume beyond this band respectively. If a patch is foreground-free, $1\%$ voxels are randomly sampled. 
 
 While using advanced U-shaped architectures can lead to marginal improvements, dynamic weighting and random sampling are indispensable. Both weighting and sampling are done during training phase while patches are fed into the network. 
\begin{figure}
	\begin{center}
			\includegraphics[width=0.25\textwidth,height=0.25\textwidth]{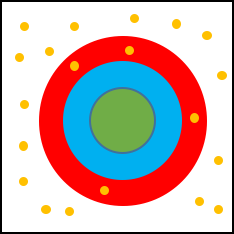} 
			\qquad
			\includegraphics[width=0.25\textwidth,height=0.25\textwidth]{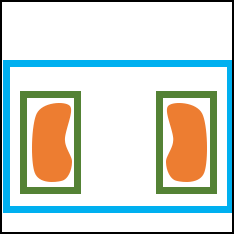} 
	\end{center}
	\caption{ \label{fig:sampling} (Left) Background sampling approach. Foreground vessel, in green, surrounded by two bands, blue and red, at distance 2 and 4 voxels. Equivalent foreground volume, 20\% and 80\%, is randomly sampled from the red band and the volume beyond this band respectively. (Right) Experiments evaluation regions. The first region is the whole region of interest defined per subject ground truth outlined in blue. The second region is the kidney bounding box outlined in green.} 
\end{figure} 
\section{Experiments}
In our experiments, we use volumetric CT-scans from 236 subjects. The average spacing is $0.757 \times 0.757 \times 0.906mm$, with standard deviation $0.075  \times 0.075 \times 0.327mm$. Kid-Net is trained using $3D$ patches from 99 cases, while 30 and 107  cases are used for validation and testing respectively. Training patches are presliced to reduce I/O bottleneck. They are uniformly sliced from CT-scans random points at background, collecting system, artery and vein centerlines. Training with foreground-free patches is mandatory. When eliminated, performance degrades because  the network learns that every patch has a foreground object, and segments accordingly, which is false.


Two training schema, with and without random sampling, are evaluated. In both schema dynamic weighting following algorithm~\ref{alg:weighting} is applied. Training without both dynamic weighting and random sampling leads to a degenerate segmentation -- background class-bias. Kid-Net is trained using Keras API~\cite{chollet2015keras}, Adam optimizer~\cite{kingma2014adam}, and a categorical cross entropy loss function. Segmentation results are evaluated using dice-coefficient -- F1 score~\cite{sorensen1948method}. 

Both artery and vein have tree-structure; they are thick near aorta and vena cava, while fine at terminals near renal artery and renal vein. These fine vessels are most difficult to annotate, i.e. most valuable to acquire. Dice-coefficient is biased against fine details in such tree-structure. To overcome such limitation, we evaluate the two regions depicted in figure~\ref{fig:sampling}. The first region is based on the ground truth region of interest. This evaluates the whole tree-structure including the thick branches-- aorta and vena cava. The second region is the kidney bounding box. It targets fine vessels in direct contact with the kidney.
 
The ground truth region of interest (ROI) is subject dependent. During evaluation, we clip our output in z-axis, based on the per subject ground truth ROI. It is worth-noting that the ground truth annotation is incomplete for two reasons. First, vessels in direct connect with kidney are challenging to annotate due to their tiny size, thus the ground truth is a discretized vessel islands -- 12 islands on average. Second, vessels far from kidney have little value, for kidney surgery, to annotate and so are typically missing. To avoid penalizing valid segmentations, we evaluate predictions overlapping with known ground-truth vessel islands. This evaluation approach reduces the chances of falsely penalizing unannotated detections. It also aligns with the premise that neural networks assist, but not replace, human especially in medical applications.

 
 	\begin{table}[H]
 	\caption{Quantitative evaluation for different training schema in two evaluation regions.  Dynamic Weighting (DW) plus Random Sampling (RS) achieves the highest accuracies. Artery F1 score is the highest.}
 	\centering
 	\begin{tabular}{| l | c |  c || c | c |}
 		\hline
 		& \multicolumn{2}{c||}{Whole ROI} & \multicolumn{2}{c|}{Kidney Bounding box}\\\hline 
 		& DW & DW+RS & DW & DW+RS\\\hline 
 		Artery   & 0.86 & \textbf{0.88} & 0.72   & \textbf{0.72}\\\hline 
 		Vein   & \textbf{0.59} & 0.57 &  0.60  & \textbf{0.67} \\\hline
 		Ureter   & 0.32 & \textbf{0.62} &  0.41 & \textbf{0.63} \\\hline
 	\end{tabular}
 	\label{tbl:quantity_eval}
 \end{table}

 Table~\ref{tbl:quantity_eval} summarizes the quantitative results and highlights our network ability to segment both coarse and fine vessels around the kidney. Artery segmentation is the most accurate because all scans are done during arterial phase. This suggests that better vein and ureter segmentations are feasible if venous and waste-out scans are available. The thick aorta boosts artery segmentation F1 score in the whole ROI. In the kidney bounding box, tiny artery vessels become more challenging, and F1 score relatively decreases. The same argument explains vein vessels F1 score. Since arterial scans are used, concealed vena cava penalizes F1 score severely in the whole ROI region. This observation manifests in figure~\ref{fig:qualitative_results}, second column. While aorta is easy to segment and boosts F1 score, vena cava is more challenging and thus F1 score degrades.
 

Among the three kidney vessels, the collecting system is the most challenging. Due to their tiny size, it is difficult to manually annotate or automatically segment. Ureter vessels ground truth annotations are available only within the kidney proximity-- far ureter are less relevant. This explains why ureter F1 score is similar in both evaluation regions. Ureter class is assigned the highest weight due to its relative small size. This leads to a lot of false ureter positives. While random sampling has limited effect on artery and vein F1 score, its merits manifest in ureter segmentation. It boosts segmentation accuracy by 30\% and 22\% in the whole ROI and kidney bounding box respectively. Thus, It is concluded that both dynamic weights and random sampling are essential to achieve accurate tiny vessels segmentation.
 
Figure~\ref{fig:qualitative_results} shows qualitative results and highlights vessels around the kidney. All CT-slices are rendered using soft tissue window--level=40, width=400. Vein and collecting system segmentation are the most challenging. The second column shows a shortcoming case due to a concealed vein.  Fine vessels near kidney are the most difficult to annotate. Manually annotating such vessels can be cumbersome and time consuming. Thats why Kid-Net is valuable; its training schema enables fine anatomy segmentation in high resolution CT-volumes, and voids GPU memory constraints. 
 \newcommand\widthOfFigures{0.24}
  \begin{figure}[t]
 	\includegraphics[width=\widthOfFigures\textwidth,height=0.25\textwidth]{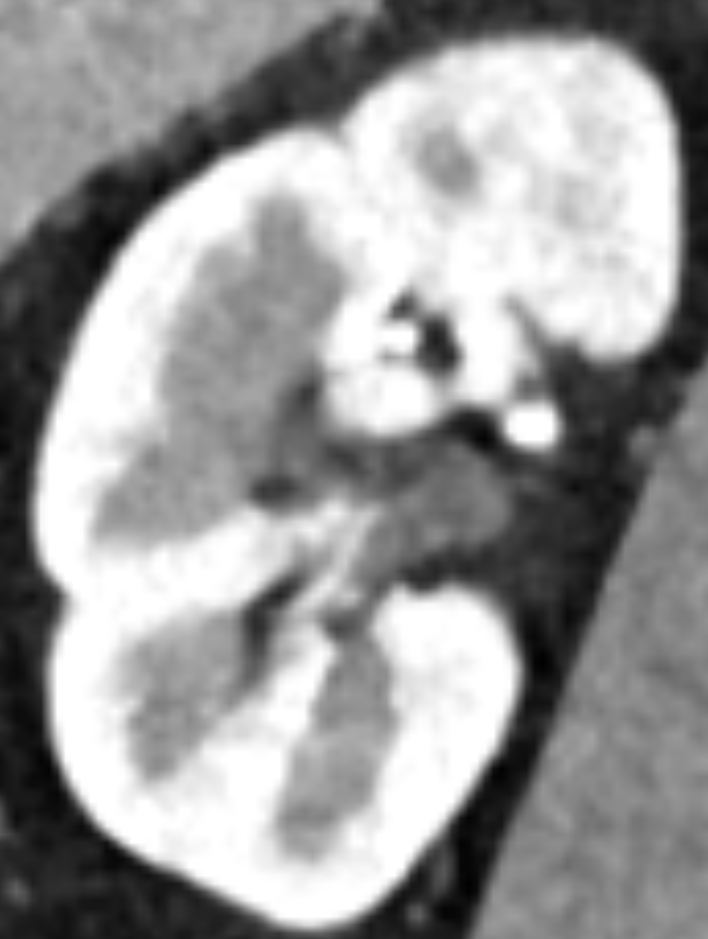}
 	\hfill\includegraphics[width=\widthOfFigures\textwidth,height=0.25\textwidth]{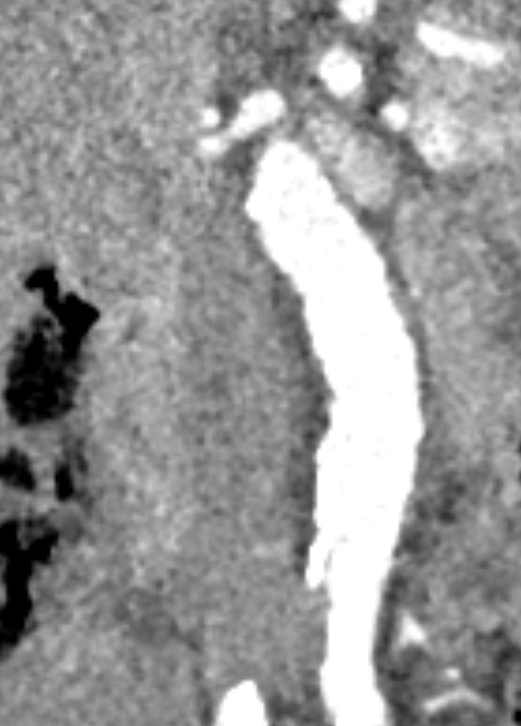}
 	\hfill\includegraphics[width=\widthOfFigures\textwidth,height=0.25\textwidth]{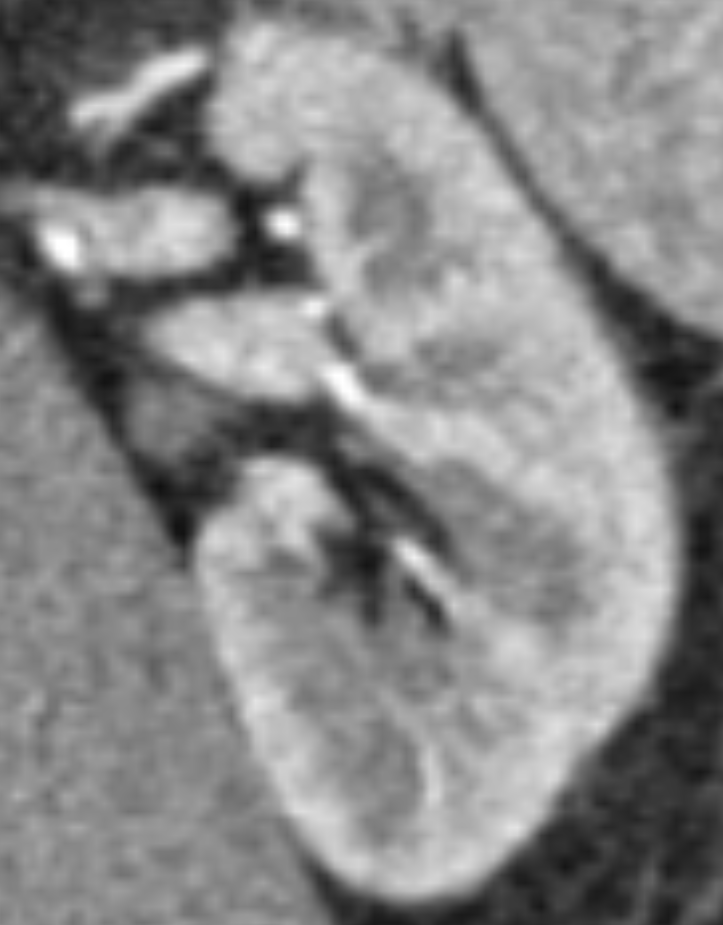}
 	\hfill\includegraphics[width=\widthOfFigures\textwidth,height=0.25\textwidth]{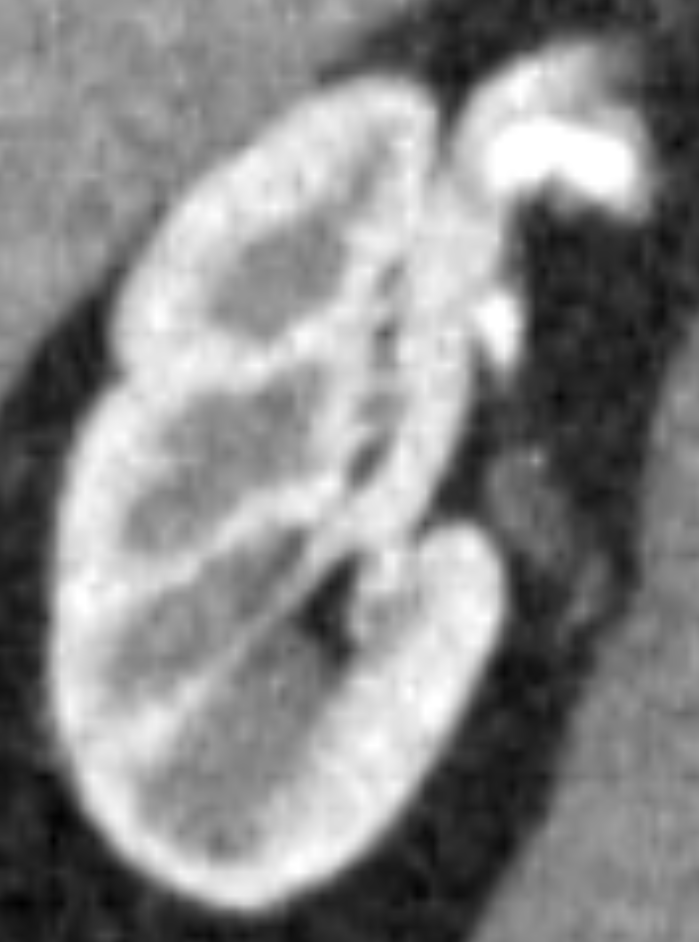}
 	
 	\includegraphics[width=\widthOfFigures\textwidth,height=0.25\textwidth]{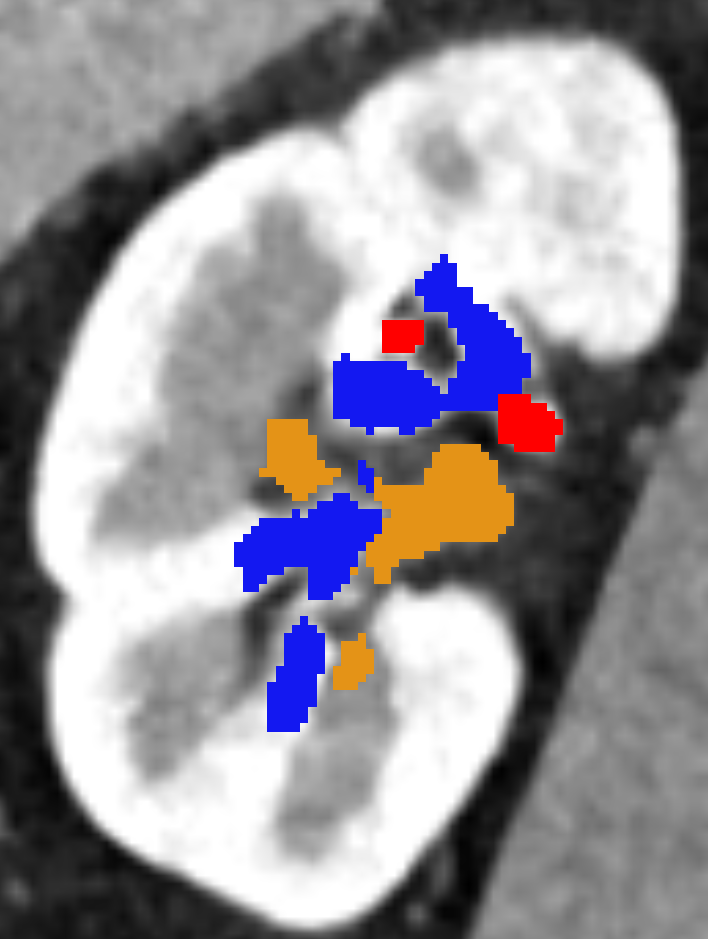}
 	\hfill\includegraphics[width=\widthOfFigures\textwidth,height=0.25\textwidth]{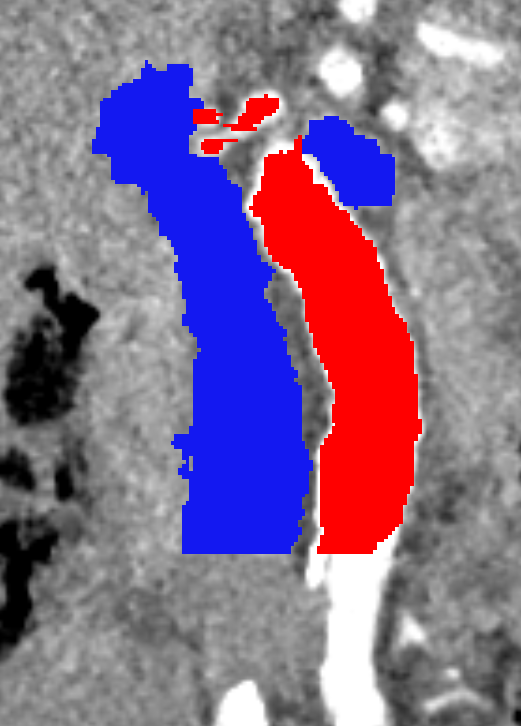}
 	\hfill\includegraphics[width=\widthOfFigures\textwidth,height=0.25\textwidth]{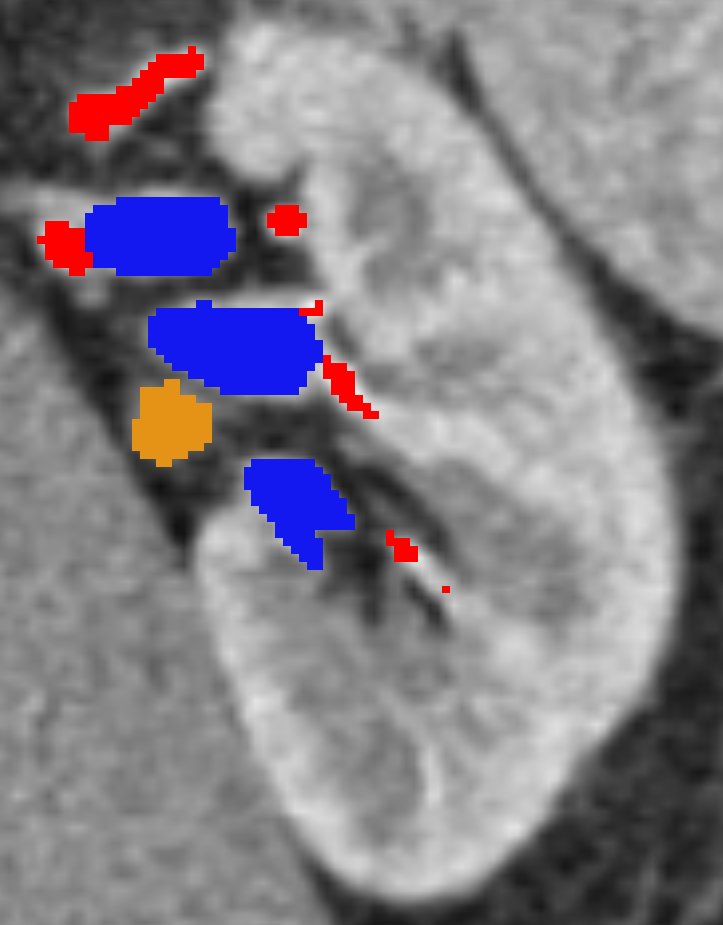}
 	\hfill\includegraphics[width=\widthOfFigures\textwidth,height=0.25\textwidth]{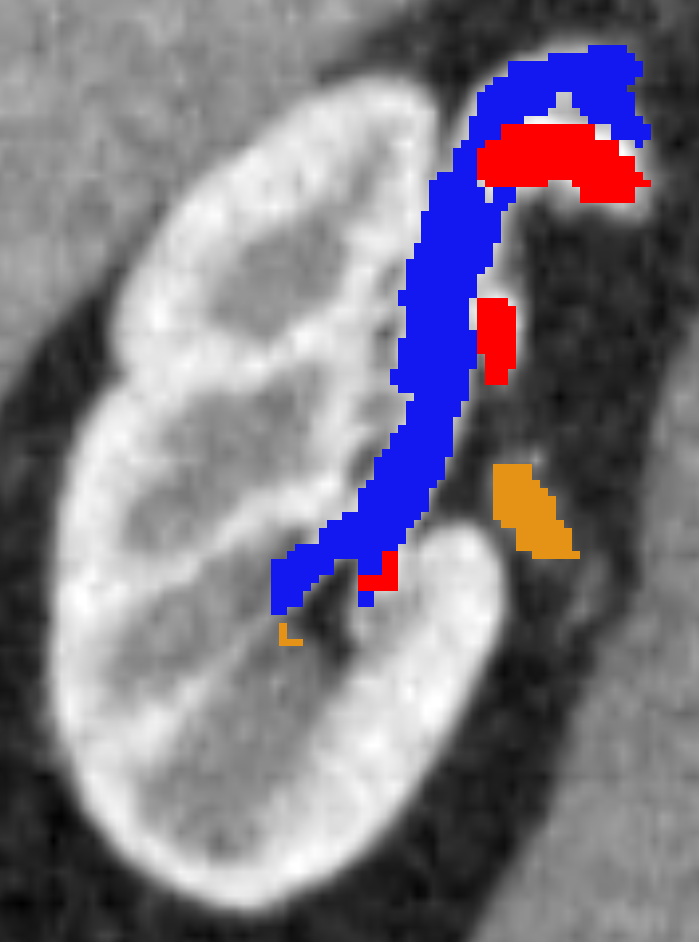}
 	
 	\includegraphics[width=\widthOfFigures\textwidth,height=0.25\textwidth]{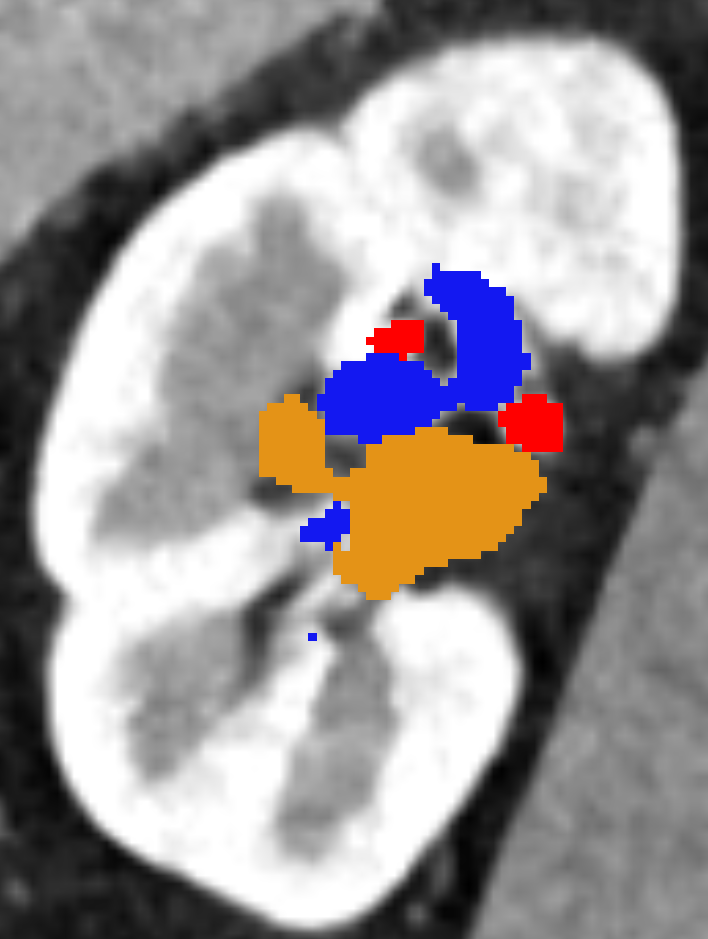}
 	\hfill\includegraphics[width=\widthOfFigures\textwidth,height=0.25\textwidth]{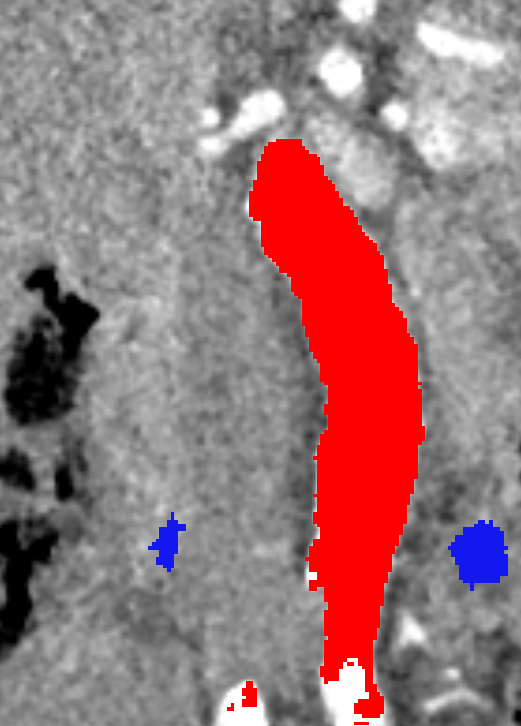}
 	\hfill\includegraphics[width=\widthOfFigures\textwidth,height=0.25\textwidth]{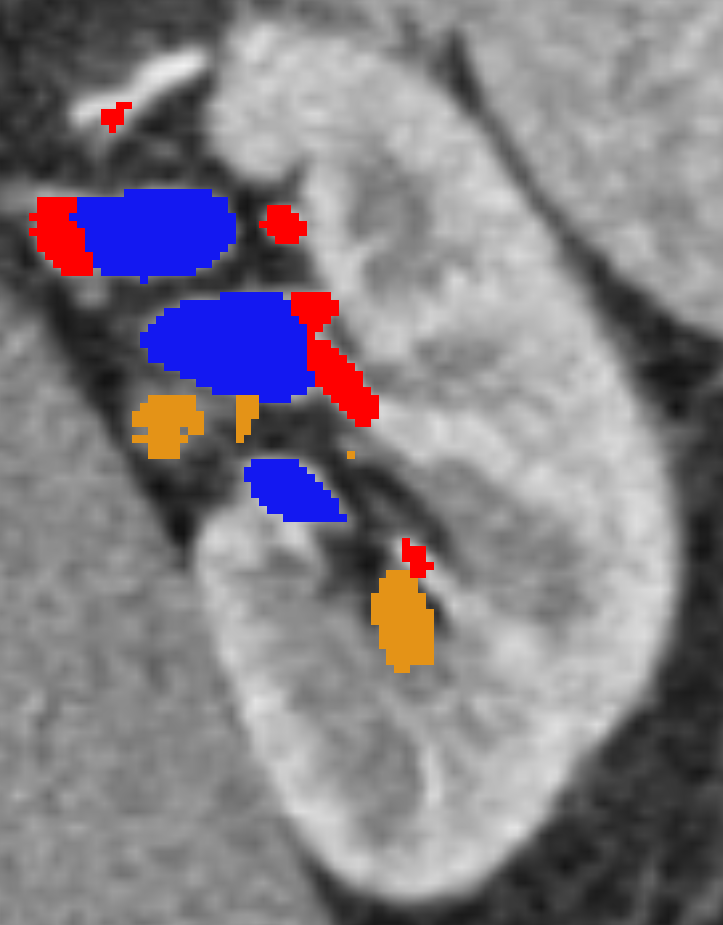}
 	\hfill\includegraphics[width=\widthOfFigures\textwidth,height=0.25\textwidth]{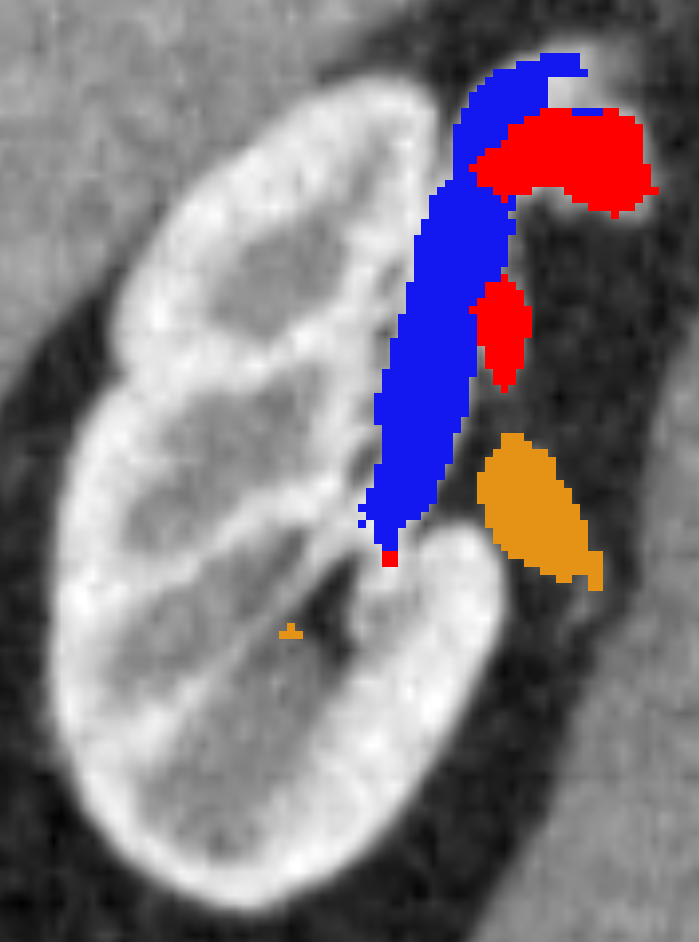}
 	\caption{Qualitative evaluation results. Rows contain raw CT slice, ground truth annotation, and Kid-Net segmentation respectively. Artery, vein, and ureter are highlighted in red, blue, and orange. Best seen in color and zoom}
 	\label{fig:qualitative_results}
 \end{figure}

\section{Conclusion}
We propose Kid-Net, a convolution neural network for kidney vessels segmentation. Kid-Net achieves great performance segmenting different vessels by processing $3D$ patches. Fitting a whole CT-volume in memory, to train a neural network, is no longer required. We propose a two-fold solution to balance foreground and background distributions in a training dataset. Dynamically weighting voxels resolves unbalanced data. Assigning higher weights to randomly sampled background voxels effectively reduces false positives. The proposed concepts are applicable to other fine segmentation tasks.  

\bibliography{report}   
\bibliographystyle{splncs03}   

\end{document}